\documentclass{article} % For LaTeX2e
\usepackage{iclr2025_conference,times}

\usepackage{amsmath,amsfonts,bm}

\def\eqref#1{equation~\ref{#1}}
\def\1{\bm{1}}

\DeclareMathAlphabet{\mathsfit}{\encodingdefault}{\sfdefault}{m}{sl}
\SetMathAlphabet{\mathsfit}{bold}{\encodingdefault}{\sfdefault}{bx}{n}

\usepackage{subcaption}
\usepackage{mathtools}
\usepackage{amssymb}
\usepackage[hidelinks]{hyperref}
\usepackage{wrapfig}
\usepackage{float}
\usepackage{url}
\usepackage{booktabs}
\usepackage{multirow}
\usepackage{arydshln}
\usepackage{caption}

\renewcommand{\eqref}[1]{Eq.~(\ref{#1})}
\newcommand{\ie}{\textit{i.e.}}
\newcommand{\eg}{\textit{e.g.}}

\newcommand{\bmeps}{{\bm{\epsilon}}}
\newcommand{\bmtheta}{{\bm{\theta}}}

\newcommand{\bmphi}{{\bm{\phi}}}

\newcommand{\bmkappa}{{\bm{\kappa}}}

\newcommand{\bmf}{{\bm{f}}}
\newcommand{\bmv}{{\bm{v}}}

\newcommand{\bmx}{{\bm{x}}}
\newcommand{\bmy}{{\bm{y}}}
\newcommand{\bmz}{{\bm{z}}}

\newcommand{\bbE}{\mathbb{E}}
\newcommand{\bbR}{\mathbb{R}}

\newcommand{\cM}{{\mathcal{M}}}
\newcommand{\cN}{{\mathcal{N}}}

\newcommand{\cT}{{\mathcal{T}}}

\DeclareMathOperator{\unif}{unif}
\DeclareMathOperator{\id}{id}
\DeclareMathOperator{\sg}{sg}

\DeclareMathOperator{\HPF}{HPF}

\DeclareMathOperator{\Proj}{Proj}

\title{Align Your Tangent: Training Better Consistency Models via Manifold-Aligned Tangents}

\author{Beomsu Kim*, Byunghee Cha*, Jong Chul Ye\\
Graduate School of AI, KAIST \\
*Equal Contribution \\
}

\iclrfinalcopy % Uncomment for camera-ready version, but NOT for submission.
\begin{document}

\maketitle

\vspace{-3mm}
\begin{abstract}
With diffusion and flow matching models achieving state-of-the-art generating performance, the interest of the community now turned to reducing the inference time without sacrificing sample quality.
Consistency Models (CMs), which are trained to be consistent on diffusion or probability flow ordinary differential equation (PF-ODE) trajectories, enable one or two-step flow or diffusion sampling.
However, CMs typically require prolonged training with large batch sizes to obtain competitive sample quality.
In this paper, we examine the training dynamics of CMs near convergence and discover that CM tangents -- CM output update directions -- are quite oscillatory, in the sense that they move parallel to the data manifold, not towards the manifold.
To mitigate oscillatory tangents, we propose a new loss function, called the {\em manifold feature distance (MFD)}, which provides manifold-aligned tangents that point toward the data manifold.
Consequently, our method -- dubbed  {\em Align Your Tangent (AYT)} -- can accelerate CM training by orders of magnitude and even out-perform the learned perceptual image patch similarity metric (LPIPS).
Furthermore, we find that our loss enables training with extremely small batch sizes without compromising sample quality.
Code: \url{https://github.com/1202kbs/AYT}
\end{abstract}

\vspace{-4mm}
\section{Introduction}
\vspace{-2mm}

Diffusion models (DM) \citep{dickstein2015thermo,ho2020ddpm,song2021ddim} and flow models (FM) \citep{liu2022rf,liu2022rfmp,lipman2023fm} have achieved remarkable progress in generative modeling over the past few years. Their strength lies in their ability to trade-off sample quality with sampling cost. Concretely, by increasing the number of score model or velocity evaluations during sample synthesis, one can reduce error in solving diffusion SDEs or flow ODEs, and thus enhance the quality of synthesized samples \citep{song2021sde,lipman2023fm}. With DMs and FMs achieving state-of-the-art generative performance \citep{dhariwal2021beat,karras2022edm,karras2023edm2,esser2024scalingrf}, the interest of the community turned to reducing the inference cost without compromising sample quality \citep{lu2022dpmsolver,dockhorn2022genie,salimans2022pd,zhang2023deis,kim2023dmcmc}.

One promising learning-based approach to accelerating DMs and FMs is consistency models (CM) \citep{song2023cm}, which are trained to transport noise to data along PF-ODE trajectories with only a minimal number of, \eg, one or two, neural net evaluations. However, CM learning is often unstable, and is prone to divergence during training \citep{song2023cm}. Subsequent works have found that better hyper-parameter choices \citep{song2024icm,lu2025scm}, techniques such as truncation \citep{lee2025tcm}, or joint learning of diffusion score or flow velocity \citep{kim2024ctm,boffi2025flow,geng2025mean,sabour2025ayf} can accelerate and stabilize training.

In this paper, we take an orthogonal, training dynamics-based approach to improving CMs by examining and enhancing CM loss functions. While there are works which propose better loss functions for DM or FM training \citep{hoogeboom2023simple,kim2025simple,lin2024perceptual,berrada2025perceptual}, losses for CMs have been left relatively under-explored after the pseudo-Huber loss gained popularity due to its ability to reduce variance during training \citep{song2024icm}. The learned perceptual image patch similarity (LPIPS) metric \citep{zhang2018lpips} has shown to be a powerful loss for training CMs \citep{song2023cm,kim2024ctm}, but construction of LPIPS involves extensive engineering such as supervised pre-training on ImageNet \citep{imagenet} and fine-tuning on a human-curated dataset of patch similarity.

Given such situation, in our work, we propose a loss function which is as powerful as LPIPS for training CMs, but is simpler to construct and does not require human supervision. Unlike pseudo-Huber or LPIPS, our loss is grounded on a rigorous analysis of CM training dynamics, such that it consists of interpretable design choices. We test our loss function on a number of variety of image generation tasks, and show that it accelerates CM training by orders of magnitude when compared to training with the pseudo-Huber loss. Furthermore, our loss simultaneously improves generative performance, and even outperforms LPIPS.

Concretely, our contributions can be summarized as:
\begin{itemize}
    \item \textbf{We discover a potential cause of slow convergence in CM training (Section~\ref{sec:analysis}).} We examine the training dynamics of CMs near convergence and identify that tangents, \ie, update directions for CM outputs, contain non-trivial amount of components which oscillate parallel to the data manifold. We hypothesize that such oscillatory components can hinder the convergence of CMs, and that one must amplify manifold-orthogonal components, \ie, components which point towards the data manifold, to enhance performance.
    \item \textbf{We propose the manifold feature distance to accelerate convergence (Section~\ref{sec:method}).} We discover that when CM loss is computed in a feature space, tangents are linear combinations of the rows of the feature map Jacobian. Inspired by this observation, we design manifold feature maps whose Jacobian consists of directions that point toward the data manifold. Consequently, computing consistency losses with our manifold feature distance provides manifold-aligned tangents with minimal oscillatory components.
    \item \textbf{We verify our method on a number of benchmark tasks (Section~\ref{sec:experiments}).} We train CMs on standard benchmark datasets CIFAR10 and ImageNet $64 \times 64$ with our manifold feature distance. We observe that our loss accelerates convergence by orders of magnitude compared to training with the pseudo-huber loss, and beats LPIPS. Furthermore, we discover that training with manifold feature distance is robust to batch size, yielding competitive FID scores with batch size as small as $16$. Overall, experiments corroborate our hypothesis that oscillatory components in tangents hinder CM convergence.
\end{itemize}

\begin{figure}[t]
\vspace{-6mm}
% \hspace{-2mm}
\centering
\includegraphics[width=0.4\linewidth]{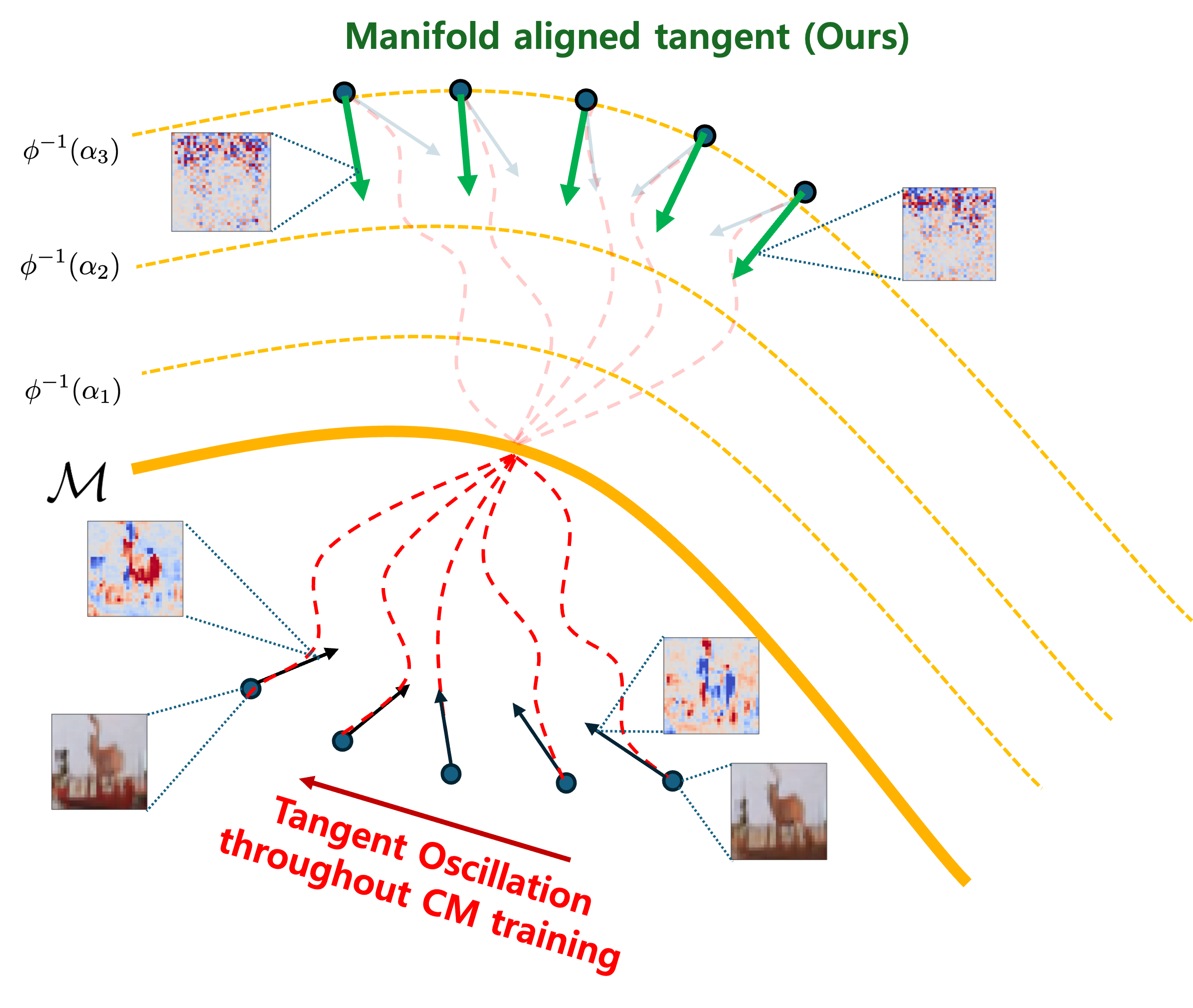}
\hspace{4mm}
\includegraphics[width=0.4\linewidth]{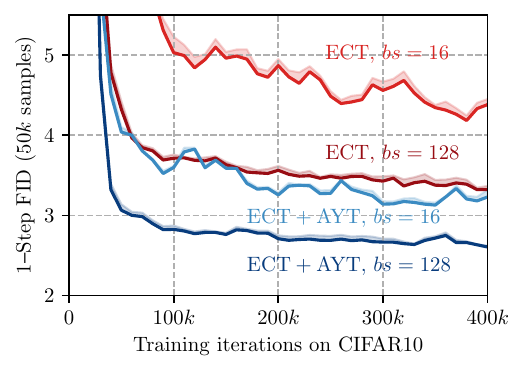}
\vspace{-2mm}
\caption{\textbf{Left:}  CM tangents, \ie, CM output update directions, exhibit large oscillations throughout training. To mitigate this, we learn feature maps $\phi$ whose level sets $\phi^{-1}(\alpha)$ model increasingly perturbed data manifolds, so feature map gradients point towards the manifold. CM tangents in the feature space are expressed as linear combinations of feature map gradients, so we obtain manifold-aligned tangents. \textbf{Right:} \textit{Manifold-aligned tangents (AYT) enable up to $\times 10$ faster convergence and competitive FIDs with $\times 1/8$ batch size (bs)}. We use Easy Consistency Training (ECT) \citep{geng2025ect}. Shaded regions indicate min/max FIDs over three sample generation trials.}
\label{fig:main}
\vspace{-5mm}
\end{figure}

\section{Related Works}

% \textbf{Regularization for GANs.} Various regularization techniques have been proposed to improve the efficiency of GAN training. Among them, one of the most extensively studied issues is the overfitting of the discriminator. \cite{mo2020freeze} alleviated this problem by applying early stopping before fine-tuning the discriminator. Subsequently, many studies \citep{zhao2021bcr,karras2020ada,hou2023augmentation,jeong2021traininggan} have introduced data augmentation to prevent overfitting and, at the same time, to enhance the robustness of the model. In addition, another line of research has focused on incorporating auxiliary tasks into the discriminator. For example, \cite{chen2019ssgan} trained the discriminator to predict the rotation angle of randomly rotated real or generated images, while \cite{kang2020contragan} further strengthened the discriminator’s representation learning by introducing contrastive learning. 

% Data augmentation for generative adversarial networks \citep{zhao2021bcr,karras2020ada,hou2023augmentation,jeong2021traininggan}. Early stopping \citep{mo2020freeze}. Patchwise similarity regularization \citep{park2020cut} for image-to-image translation. Self-supervised loss \citep{chen2019ssgan,kang2020contragan,hou2021ssganla}.

% \textbf{Study of generalization in denoisers.} Generalization in diffusion models \citep{kadkhodaie2024gahb,kamb2025creativity}.

\textbf{Regularization for DMs and CMs.}  Similar to GANs, regularization for efficient training has also been actively explored in diffusion and consistency models. In particular, early stopping has been introduced to mitigate overfitting, which often occurs at small timesteps~\citep{nichol2021iddpm,lee2025tcm}. Although simple, this approach is effective in preventing the model from over-adapting to the data distribution and has therefore received significant attention. Subsequently, various data augmentation strategies have been proposed. For example, non-leaking augmentation~\citep{karras2022edm} and noise perturbation techniques~\citep{daras2025omni,ning2023input} have been employed to improve generalization and enhance robustness under diverse conditions. More recently, research has shifted toward incorporating more sophisticated auxiliary learning signals. For instance, contrastive learning objectives have been adopted to encourage the model to acquire more discriminative image samples~\citep{stoica2025contrastive}, while the outputs of pre-trained representation models have been aligned with intermediate diffusion features to accelerate training and improve convergence stability~\citep{yu2025repa,jeong2025track,chefer2025videojam}.

% Data augmentation \citep{karras2022edm,daras2025omni}. Early stopping for DMs \citep{nichol2021iddpm} and CMs \citep{lee2025tcm}. Auxiliary losses such as patch-wise regularization \citep{kim2024unsb}, representation alignment \citep{yu2025repa,jeong2025track,chefer2025videojam}. Input perturbation robustness for mitigating exposure bias \citep{ning2023input}. Contrastive flow matching \citep{stoica2025contrastive}. Smoothing via EMA \citep{song2020improved}. Dropout \citep{geng2025ect}. Generator-Augmented Flow \citep{issenhuth2025improving} Augmentation Equivariance \citep{zhou2025afldm}

% \textbf{Metric learning.} Deep metric learning with triplet network \citep{hoffer2014metric}. Learned perceptual metric \citep{zhang2018lpips}.

% \textbf{Gradient surgery.} Gradient surgery for multi-task learning \citep{yu2020surgery}. Used in context of accelerating DM training \citep{go2023negative,hang2023minsnr}.

% \textbf{Corruption-based generation.} Soft diffusion \citep{daras2023soft} and cold diffusion \citep{bansal2023cold}. Ambient diffusion \citep{daras2023ambient,daras2025omni}.

\textbf{Perceptual objectives.} Various studies have explored the use of perceptual metrics to facilitate the training of diffusion and consistency models. However, due to the nature of score matching, directly minimizing perceptual metrics can adversely affect the training of diffusion models. To address this, some methods use perceptual losses only after the diffusion model has been pretrained \citep{lin2024perceptual}, while others incorporate them as auxiliary losses during training \citep{berrada2025perceptual}. In the case of consistency models, perceptual metrics such as LPIPS \citep{zhang2018lpips} can be directly employed as consistency losses without compromising theoretical guarantees \citep{song2023cm}.

% \textbf{Manifold informed learning.} Manifold informed guided diffusion \citep{he2024manifold}. Manifold tangent classifier \citep{rifai2011manifold}.

\textbf{Fast sampling of diffusion and flow models.} While diffusion models have demonstrated remarkable performance in image and video generation, their sampling process often requires hundreds to thousands of steps, resulting in significant computational cost. To address this limitation, a wide range of approaches have been proposed to enable fast sampling, where high-quality samples can be generated with only a few steps. Early studies \citep{lu2022dpmsolver,zhang2023deis,dockhorn2022genie,zhou2024amed} primarily focused on improving ODE solvers. By mitigating error accumulation across timesteps, these methods reduced the required number of sampling steps to about 10. Beyond solver improvements, several approaches have aimed to directly train models capable of efficient sampling. A representative example is Rectified Flow \citep{liu2022rf,liu2022rfmp,liu2024slimflow,liu2024instaflow,lee2024rfpp,kim2025simple}, which straightens the ODE trajectory from noise to image, thereby minimizing error accumulation under a small number of steps. Another line of research is diffusion model distillation \citep{salimans2022pd,meng2023distill,kim2024dode}. In this paradigm, a pretrained diffusion model is distilled into a new single-step generative model by leveraging objectives such as diffusion losses. More recently, flow map–based approaches \citep{song2023cm,kim2024ctm,kim2025gctm,sabour2025ayf,geng2025mean} have been introduced, which learn the trajectory of an ODE directly, predicting the destination at a target timestep from an input at a source timestep. These methods are particularly notable as they can be applied both in the context of distillation and from-scratch training. Consistency models can be interpreted as a special case of this family, where the target timestep is set to zero, allowing the model to predict the final image directly from noise.

% ReFlow \citep{liu2022rf,liu2022rfmp,liu2024slimflow,liu2024instaflow,lee2024rfpp,kim2025simple}. Distillation \citep{salimans2022pd,meng2023distill,kim2024dode}. Fast solvers \citep{lu2022dpmsolver,zhang2023deis,dockhorn2022genie,zhou2024amed}. Flow maps \citep{song2023cm,kim2024ctm,kim2025gctm,sabour2025ayf, Geng2025mean}. Combining with adversarial learning \citep{kong2024act}.

\vspace{-2mm}
\section{Background}
\vspace{-2mm}

Our goal is to learn a generative model of a data distribution $p(\bmx)$ supported on $\bbR^d$. Given a forward or corruption process from data $\bmx \sim p(\bmx)$ to noise $\bmeps \sim \cN(\bm{0},\bm{I})$
\begin{align}
    \textstyle \bmx_t = \alpha_t \bmx + \sigma_t \bmeps \label{eq:forward}
\end{align}
parametrized by time $t \in [0,\infty)$ with boundary conditions $\bmx_0 = \bmx$ and $\lim_{t \rightarrow \infty} \alpha_t/\sigma_t = 0$, let us denote the distribution of $\bmx_t$ at time $t$ as $p_t$. The corresponding probability flow ordinary differential equation (PF-ODE), \ie, an ODE whose marginal equals $p_t$ for all time $t$, is given by
\begin{align}
    \textstyle d\bmx_t = \bbE_{\bmx,\bmeps|\bmx_t}[\dot{\bmx}_t] \, dt = \bbE_{\bmx,\bmeps|\bmx_t}[\dot{\alpha}_t \bmx + \dot{\sigma}_t \bmeps] \, dt \label{eq:pfode}
\end{align}
and velocity can be learned by solving flow matching \citep{lipman2023fm,albergo2023stochastic}
\begin{align}
    \textstyle \min_{\bmv} \bbE_{\bmx,\bmeps,t} [\| (\dot{\alpha}_t \bmx + \dot{\sigma}_t \bmeps) - \bmv(\bmx_t,t) \|_2^2].
\end{align}
In particular, we note that $p_0 = p$ and $p_T \approx \cN(\bm{0},\sigma_T^2\bm{I})$ for a sufficiently large time $T$, so we can sample from $p$ by sampling $\bmeps \sim \cN(\bm{0},\bm{I})$ and solving the PF-ODE down from time $t = T$ to $0$ with terminal condition $\bmx_T = \sigma_T \bmeps$. However, numerical integration of the PF-ODE involves multiple evaluations of the velocity, so the generation process is often slow and costly.

A consistency model (CM) $\bmf_\bmtheta : \bbR^d \times [0,\infty) \rightarrow \bbR^d$ with boundary condition $\bmf_\bmtheta(\cdot,0) = \id_{\bbR^d}$ is trained to be consistent, \ie, to have identical outputs, on PF-ODE trajectories \citep{song2023cm,song2024icm,lu2025scm}. Hence, an optimal CM $\bmf_{\bmtheta^*}$ will map all points on the PF-ODE trajectory back to its initial point at $t = 0$ with a single function evaluation. In particular, the output $\bmf_{\bmtheta^*}(\sigma_T\bmeps,T)$ for $\bmeps \sim \cN(\bm{0},\bm{I})$ will be distributed according to $p$, so a CM can bypass the computational burden of solving the PF-ODE.

The discrete CM objective \citep{song2023cm} forces $\bmf_\bmtheta$ to be consistent on consecutive timesteps\footnote{While the original CM objective also contains a time-dependent weight function $w(t)$, we omit it without loss of generality since it can be absorbed into the density function for $t$.}:
\begin{align}
    \textstyle \min_\bmtheta \bbE_{\bmx,\bmeps,t,\Delta t}[(\Delta t)^{-1} d(\bmf_\bmtheta(\bmx_t,t),\bmf_{\sg[\bmtheta]}(\bmx_{t-\Delta t},t-\Delta t))] \label{eq:disc_cm}
\end{align}
Here, $d$ is a loss function such as LPIPS, mean squared error, and pseudo-Huber. With the choice of $d(\bmx,\bmy) = \frac{1}{2} \|\bmx - \bmy\|_2^2$, one can derive an alternative objective with equivalent gradients:
\begin{gather}
    \textstyle \min_\bmtheta \bbE_{\bmx,\bmeps,t,\Delta t}[ \bmf_\bmtheta(\bmx_t,t)^\top (\Delta \bmf_{\sg[\bmtheta]}(\bmx_t,t)/\Delta t)], \label{eq:disc_cm_alt} \\
    \Delta \bmf_{\sg[\bmtheta]}(\bmx_t,t)/\Delta t \coloneqq (\bmf_{\sg[\bmtheta]}(\bmx_t,t) - \bmf_{\sg[\bmtheta]}(\bmx_{t-\Delta t},t-\Delta t))/\Delta t. \label{eq:disc_tangent}
\end{gather}
Depending on how we approximate $\bmx_{t-\Delta t}$ given $\bmx_t = \alpha_t \bmx + \sigma_t \bmeps$ in \eqref{eq:disc_tangent}, we obtain consistency distillation (CD) and consistency training (CT): the former uses $\bmx_{t-\Delta t} \approx \bmx_t - \bmv(\bmx_t,t) \cdot \Delta t$ to distill the velocity, whereas the latter uses $\bmx_{t-\Delta t} \approx \bmx_t - \dot{\bmx}_t \cdot \Delta t \approx \alpha_{t-\Delta t} \bmx + \sigma_{t-\Delta t} \bmeps$. 

Letting $\Delta t \rightarrow 0$ in \eqref{eq:disc_cm_alt} yields the continuous CM objective \citep{lu2025scm}
\begin{gather}
    \textstyle \min_\bmtheta \bbE_{\bmx,\bmeps,t}[\bmf_\bmtheta(\bmx_t,t)^\top (d\bmf_{\sg[\bmtheta]}(\bmx_t,t)/dt)], \label{eq:contin_cm} \\
    \textstyle d\bmf_{\sg[\bmtheta]}(\bmx_t,t)/dt = \nabla_{\bmx_t} \bmf_{\sg[\bmtheta]}(\bmx_t,t) (d\bmx_t/dt) + \partial \bmf_{\sg[\bmtheta]}(\bmx_t,t) / \partial t. \label{eq:tangent}
\end{gather}
The derivative in \eqref{eq:tangent} is called the \textit{tangent}, since it is tangential to the trajectory traced out by the CM output $\bmf_\bmtheta(\bmx_t,t)$ as $\bmx_t$ follows the PF-ODE in \eqref{eq:pfode}. Analogous to discrete CMs, we obtain continuous CD or CT depending on how we estimate $d\bmx_t/dt$ in the tangent. CD uses the flow velocity $d\bmx_t/dt = \bmv(\bmx_t,t)$, whereas CT uses $d\bmx_t/dt = \dot{\bmx}_t = \dot{\alpha}_t \bmx + \dot{\sigma}_t \bmeps$.

From \eqref{eq:disc_cm_alt} and \eqref{eq:contin_cm}, we can interpret discrete CM and continuous CM learning from an unified perspective of contracting each path $\{\bmf_{\bmtheta}(\bmx_s,s) : d\bmx_s = \bmv(\bmx_s,s)\,ds, \ \bmx_0 = \bmx\}$ along the negative tangent towards $\bmf_\bmtheta(\bmx_0,0) = \bmx_0 = \bmx$. The only difference between discrete CM and continuous CM lies in whether we calculate the tangent using finite differences or the exact derivative. Hence, we may use the tangent, both discrete and continuous, to analyze training dynamics of CMs.

\begin{figure}[t]
\centering
\includegraphics[width=1.0\linewidth]{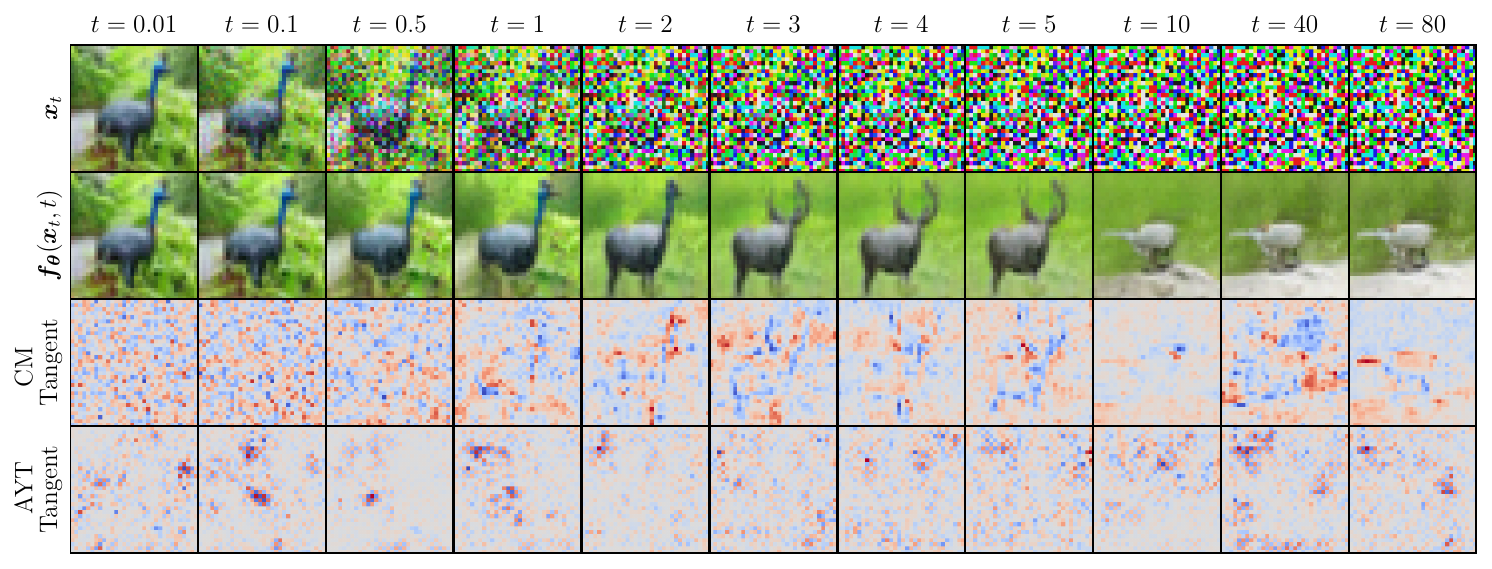}
\caption{CM tangent visualization on CIFAR10 after training to near-convergence ($400k$ iterations). \textbf{First row:} inputs $\bmx_t = \bmx_0 + t \bmeps$. \textbf{Second row:} outputs $\bmf_\bmtheta(\bmx_t,t)$. \textbf{Third row:} vanilla CM tangents computed with \eqref{eq:disc_tangent}. Tangents are averaged along the channel dimension for visualization, and red and blue pixels indicate positive and negative values, resp. \textbf{Fourth row:} manifold-aligned tangents (AYT) computed with \eqref{eq:feature_tangent}.}
\label{fig:tangent_cifar10}
\vspace{-5mm}
\end{figure}

\section{The Oscillatory Tangent Hypothesis} \label{sec:analysis}

We now further assume data is supported on a low-dimensional manifold $\cM$ in $\bbR^d$.\footnote{An \textit{$m$-dimensional manifold in $\bbR^d$} is a space in $\bbR^d$ which resembles an $m$-dimensional Euclidean space at a neighborhood of each point, and a \textit{low-dimensional manifold} means a manifold with $m \ll d$. The space which $\cM$ exists in, $\bbR^d$ in this case, is called the \textit{ambient space}. The \textit{tangent space of a manifold at a point} can intuitively be interpreted as a linear approximation of the manifold at the point \citep{lee2012manifold}. Our work assumes that the \textit{manifold hypothesis}, which asserts that high-dimensional data lie in vicinity of a low-dimensional manifold \citep{narayanan2010manifold}, holds in practice.}
Since tangents represent instantaneous changes in path $\{\bmf_{\bmtheta}(\bmx_s,s) : d\bmx_s = \bmv(\bmx_s,s)\,ds, \ \bmx_0 = \bmx\}$, small perturbations of the path can induce large variations in the tangent.
Given the stochasticity within CM training, we hypothesized that tangents are oscillatory and unlikely to guide the CM output exactly towards the low-dimensional data manifold $\cM$.
We also hypothesized that this phenomenon actually occurs in practice, and adversely affects CM convergence.
From here on, these claims will be referred to as the \textit{oscillatory tangent hypothesis}.
To validate this hypothesis, we began by examining CM tangents on CIFAR10 \citep{cifar10}.

% \textbf{Training and evaluation setting.}
% We adopted the easy consistency training (ECT) \citep{geng2025ect} setting due to its simplicity and accessibility. The only difference is that we trained models with batch size $64$ instead of $128$. We used the forward process $\bmx_t = \bmx + t \bmeps$, which corresponds to the choice of $\alpha_t = 1$ and $\sigma_t = t$ in \eqref{eq:forward}.
% Hence, we will refer to $t$ as \textit{time} or \textit{noise level}.
% Note that since ECT is a discrete CT framework, we computed tangents via \eqref{eq:disc_tangent}.
% Models were evaluated using the Frech\'{e}t Inception Distance (FID) \citep{heusel2017fid} between $50k$ generated images and all available dataset images. We reported the minimum FID score out of three random trials with fixed random seed, following standard evaluation protocol \citep{karras2022edm}.

\begin{figure}[t]
\centering
\begin{subfigure}{0.48\linewidth}
\includegraphics[width=1.0\linewidth]{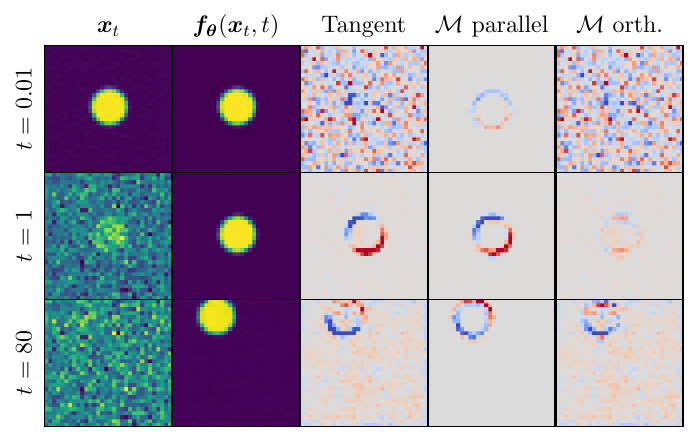}
\caption{Vanilla CM Tangent Analysis}
\end{subfigure} \hfill
\begin{subfigure}{0.48\linewidth}
\includegraphics[width=1.0\linewidth]{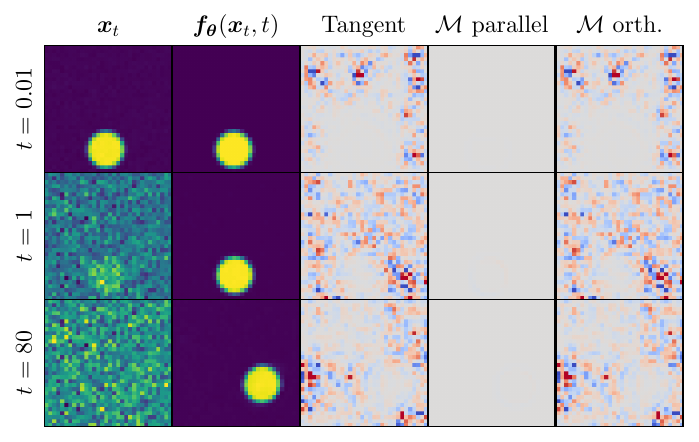}
\caption{AYT (Ours) Tangent Analysis}
\end{subfigure}
\vspace{-2mm}
\caption{Tangent analysis on 2D discs after training to near-convergence ($200k$ iterations) for vanilla CM and align your tangent (AYT). In each figure, we visualize CM inputs, CM outputs, CM tangents, manifold-parallel component of tangents, and manifold-orthogonal component of tangents.}
\label{fig:discs_analysis_viz}
\vspace{-6mm}
\end{figure}

\textbf{Observations on CIFAR10.} On CIFAR10, we optimized a CM via consistency training (CT) for $400k$ iterations until near-convergence, so there were no longer large changes in the FID score.\footnote{ECT attains 2-step FID scores of $2.20$ at iteration $100k$, and $2.11$ at iteration $400k$ \citep{geng2025ect}.} We then computed tangents at various noise levels ranging from $t = 0.01$ to $80$.
Upon visual inspection of CM tangents in the third row of Fig.~\ref{fig:tangent_cifar10}, we noticed that \textit{tangents contained structured patterns that could imply large movements along the manifold, not toward the manifold, in accordance with our hypothesis.}
To provide further evidence for the oscillatory tangent hypothesis, we performed an additional experiment on a synthetic dataset with known manifold structure.

\begin{wrapfigure}{r}{0.5\linewidth}
\vspace{-5mm}
\centering
\includegraphics[width=1.0\linewidth]{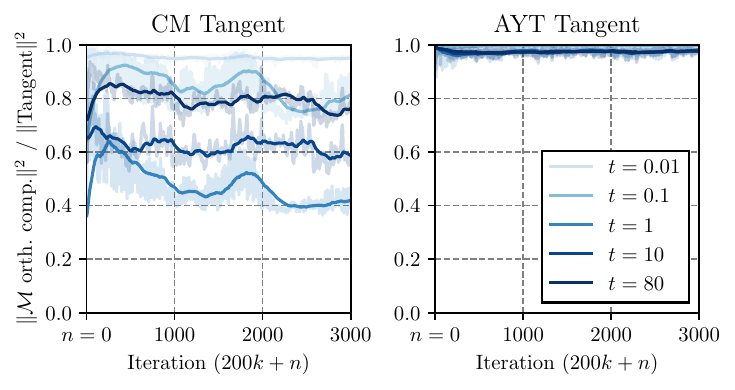}
\vspace{-7mm}
\caption{Amount of manifold-orthogonal components in tangents for vanilla CM and our manifold aligned tangents (AYT) throughout training.}
\label{fig:discs_analysis}
\vspace{-6mm}
\end{wrapfigure}

\textbf{Analysis on synthetic data.} To further analyze training dynamics of CMs, we considered the dataset of images of two-dimensional discs which move vertically or horizontally. As previously noted by \citet{kadkhodaie2024gahb}, this dataset is a two-dimensional curved manifold with tangent space at a point spanned by deformations corresponding to vertical or horizontal movement.
% We also remark that the dimension of this dataset is very small compared to the dimension of the ambient space, which is $32\times32\times3$.
% A disc image and its tangent space basis vectors are shown in Fig.~\ref{fig:discs_viz}.
Analogous to the previous experiment on CIFAR10, we trained a CM for $200k$ iterations until convergence, and computed tangents at $t \in \{0.01,0.1,1,10,80\}$ with models at iterations $\geq 200k$.

Motivated by our observations on CIFAR10, we decomposed each tangent into manifold-parallel and orthogonal components.
Concretely, given a CM output $\bmf_\bmtheta(\bmx_t,t)$, we computed its projection onto the manifold $\cM$ and the tangent space of $\cM$ at the projected point.
Let us denote the CM output, its projection, and the tangent space at the projected point as $\bmz$, $\hat{\bmz}$, and $T_{\hat{\bmz}} \cM$.
Since vectors in $T_{\hat{\bmz}} \cM$ lie along $\cM$, we defined the manifold-parallel component of the tangent $d\bmf_{\bmtheta}(\bmx_t,t)/dt$ as $\Proj_{T_{\hat{\bmz}} \cM}(d\bmf_{\bmtheta}(\bmx_t,t)/dt)$, and the manifold-orthogonal component as the remainder obtained by subtracting the manifold-parallel component from the tangent.
We note that by definition, manifold-parallel and orthogonal components are mutually orthogonal vectors.

Left panel of Fig.~\ref{fig:discs_analysis_viz} displays CM tangents and their decomposition into manifold-parallel and orthogonal components. Indeed, we see that tangents are quite oscillatory despite the CM FID having converged -- tangents contain non-trivial manifold-parallel components, especially at $t \geq 1$.
This may be concerning, because oscillatory tangents at large $t$ are at odds with the objective of CM, which is to map pure noise at large $t$ to data.
In fact, as shown in the left panel of Fig.~\ref{fig:discs_analysis}, we found overwhelmingly large amount of manifold-parallel components in the tangent. Altogether, there were strong evidences which corroborated the oscillatory tangent hypothesis. This motivated us to design a loss function which amplifies manifold-relevant components for CM training.

\section{Tangent Alignment for Consistency Model Training} \label{sec:method}

\subsection{Consistency Model Training with Feature Distance}

\citet{kim2025simple} demonstrated that using loss functions of the form $\ell(\bmx,\bmy) = \|\bmphi(\bmx) - \bmphi(\bmy)\|_2^2$ with an invertible linear map $\bmphi$ for flow matching can accelerate flow model convergence by amplifying certain directions in the model gradient. For instance, the loss function with $\bmphi = \bm{I} + \lambda \cdot \HPF$, where $\HPF$ is a high-pass filter, magnifies gradient components in the high-frequency regime by a factor of $\lambda+1$. Taking inspiration from this observation, we adopt a similar approach for designing a new loss function for training consistency models (CMs).

Let us consider using a (not necessarily linear) feature map $\bmphi : \bbR^d \rightarrow \bbR^n$ which maps points in $\bbR^d$ into a feature space $\bbR^n$ to define a feature distance $d_{\bmphi}(\bmx,\bmy) \coloneqq \| \bmphi(\bmx) - \bmphi(\bmy) \|_2$. With the squared feature distance as the CM loss function, the continuous CM objective gradient becomes
\begin{align}
& \textstyle \lim_{\Delta t \rightarrow 0} \nabla_{\bmtheta} \frac{1}{2 \Delta t} \|\bmphi(\bmf_\bmtheta(\bmx_t,t)) - \bmphi(\bmf_{\sg[\bmtheta]}(\bmx_{t-\Delta t},t - \Delta t))\|_2^2 \\
= & \textstyle \lim_{\Delta t \rightarrow 0} \frac{1}{\Delta t} (\bmphi(\bmf_\bmtheta(\bmx_t,t)) - \bmphi(\bmf_{\bmtheta}(\bmx_{t-\Delta t},t - \Delta t)))^\top \nabla_{\bmtheta} \bmphi(\bmf_\bmtheta(\bmx_t,t)) \\
= & \textstyle \lim_{\Delta t \rightarrow 0} \frac{1}{\Delta t} (\bmphi(\bmf_\bmtheta(\bmx_t,t)) - \bmphi(\bmf_{\bmtheta}(\bmx_{t-\Delta t},t - \Delta t)))^\top \bm{J}_{\bmphi}(\bmf_\bmtheta(\bmx_t,t)) \nabla_{\bmtheta} \bmf_\bmtheta(\bmx_t,t) \\
= & (d\bmphi(\bmf_\bmtheta(\bmx_t,t))/dt)^\top \bm{J}_{\bmphi}(\bmf_\bmtheta(\bmx_t,t)) \nabla_{\bmtheta} \bmf_\bmtheta(\bmx_t,t)
\end{align}
such that a CM objective with same gradient is given as
\begin{gather}
    \textstyle \min_{\bmtheta} \bbE_{\bmx,\bmeps,t} [\bmf_\bmtheta(\bmx_t,t)^\top \sg[(d\bmphi(\bmf_{\bmtheta}(\bmx_t,t))/dt)^\top \bm{J}_{\bmphi}(\bmf_{\bmtheta}(\bmx_t,t))]] \\
    \textstyle (d\bmphi(\bmf_{\bmtheta}(\bmx_t,t))/dt)^\top \bm{J}_{\bmphi}(\bmf_{\bmtheta}(\bmx_t,t)) = \sum_{i = 1}^n (d\phi_i(\bmf_{\bmtheta}(\bmx_t,t))/dt) \nabla_{\bmf_{\bmtheta}} \phi_i(\bmf_{\bmtheta}(\bmx_t,t)) \label{eq:feature_tangent}
\end{gather}
where $\bm{J}_{\bmphi}(\bmf_{\bmtheta}(\bmx_t,t))$ is the Jacobian of $\bmphi$ w.r.t. $\bmf_{\bmtheta}(\bmx_t,t)$.

It follows that when we use the squared feature distance as the loss, the $d$-dimensional vector \eqref{eq:feature_tangent} plays the role of CM tangent during optimization.
We observe that \eqref{eq:feature_tangent} is a linear combination of the rows of the Jacobian of $\bmphi$, so $\bmphi$ completely determines which direction the tangent points to.
Thus, with a judiciously chosen $\bmphi$, one can potentially suppress oscillatory components in the tangent.
However, when $\bmphi = \id_{\bbR^d}$,  which is the case of the original CM,
the Jacobian becomes the full-rank identity matrix $\bm{I}_d$, so the tangent is computed as a linear combination of the standard basis, and is free to point in any direction.
It turns out that, to align tangents toward the data manifold, one should use manifold features, which we present in the next section.

\subsection{Align Your Tangent (AYT) with Manifold Features} \label{sec:ayt}

\eqref{eq:feature_tangent} along with our observations in Section~\ref{sec:analysis} implies that an ideal feature map $\bmphi$ for optimizing CMs should possess Jacobians whose rows, \ie, gradients $\nabla_{\bmz} \phi_i(\bmz)$ for $i = 1, \ldots, n$ point toward the data manifold $\cM$. To this end, we consider $\bmphi$ such that for each coordinate $i$, its level set at zero $\phi_i^{-1}(0) = \cM$, and $\phi_i^{-1}(\alpha)$ for increasing values of $|\alpha|$ correspond to increasingly perturbed versions of $\cM$.\footnote{By $\phi_i^{-1}(\alpha)$, we mean the level set of $\phi_i$ at $\alpha$, \ie, $\phi_i^{-1}(\alpha) \coloneqq \{\bmx \in \bbR^d : \phi_i(\bmx) = \alpha \}$.} Since the gradient of a scalar-valued function is orthogonal to its level set, we can expect $\nabla_{\bmz} \phi_i(\bmz)$ would also point towards $\cM$, depending on how the manifold is perturbed. Hence, we shall call each $\phi_i$ a \textit{manifold feature}, and $d_{\bmphi}$ as a \textit{manifold feature distance}.
% We now discuss some choices of manifold features, and how to learn them.

In our work, we consider pointwise manifold perturbations of the form $\cT_\alpha \cM \coloneqq \{\cT_\alpha \bmx : \bmx \in \cM\}$, where $\cT_\alpha : \bbR^d \rightarrow \bbR^d$ is a transformation smoothly parametrized by $\alpha \in \bbR$ with $\cT_0 = \id_{\bbR^d}$. Given a collection of $n$ such transformations $\{\cT^i\}_{i=1}^n$, we can parametrize $\bmphi$ with a neural net and optimize
\begin{align}
    \textstyle \min_{\bmphi} \bbE_{\bmx,i \in [n],\alpha \in \bbR} [ \|\phi_i(\cT^i_\alpha(\bmx)) - \alpha\|_2^2] \label{eq:mf_loss}
\end{align}
such that $\phi_i(\bmx) = \alpha$ for $\bmx \in \cT^i_\alpha \cM$. In particular, with optimal $\bmphi$, $\phi_i(\bmx) = 0$ for all $\bmx \in \cM$ due to the condition $\cT_0 = \id_{\bbR^d}$.
We also remark that while isotropic perturbation of $\cM$ via, \eg, Gaussian noise addition may be sufficient to generate manifold-orthogonal feature gradients, it can also be beneficial to use anisotropic transformations to further emphasize certain off-manifold directions.

% We also remark that the set of transformations should be large enough so the tangent is sufficiently expressive.
% For instance, if one solely uses transformations which push samples off $\cM$, the gradients $\nabla_{\bmz} \phi_i(\bmz)$ will only point towards or away from $\cM$.
% Then, the CM will be penalized solely for producing off-manifold samples without regards to sample diversity, and this may cause issues similar to mode collapse in generative adversarial networks \citep{goodfellow2014gan,arjovsky2017wgan}.

\textbf{Transformation and implementation details.} We further limit our scope to the image domain, and consider image transformations for $\cT$. We consider three image degradations given by Gaussian noise perturbation, Gaussian blur, and Mixup \citep{zhang2017mixup}, four geometric transformations given by isotropic scaling, anisotropic scaling, fractional rotation, and fractional translation, and four color transformations given by perturbations in brightness, contrast, hue, and saturation. This yields a feature space of dimension $n = 15$. Readers are referred to Appendix~\ref{append:settings} for a comprehensive description of how the transformations are defined. Manifold feature $\bmphi$ is parametrized with a VGG16 classification network \citep{simonyan2015vgg}, and in the spirit of LPIPS \citep{zhang2018lpips}, we also use intermediate max-pooling features as manifold features.

\textit{Example for Gaussian Blur.} The transformation is given as $\cT_{\alpha}(\bmx) \coloneqq \bmkappa_{\alpha} \circledast \bmx$, where $\bmkappa_\alpha$ is a blurring kernel with standard deviation $\alpha$. Manifold feature can be learned by optimizing
\begin{align}
    \textstyle \min_\phi \bbE_{\bmx,\alpha \sim \unif(0,\alpha_{\max})}[( \phi(\bmkappa_\alpha \circledast \bmx) - \alpha )^2]
\end{align}
where $\unif(0,\alpha_{\max})$ is a uniform distribution on $[0,\alpha_{\max}]$, and discrete CM optimization by
\begin{align}
    \textstyle \min_\bmtheta \bbE_{\bmx,\bmeps,t,\Delta t}[(\Delta t)^{-1} (\phi(\bmf_\bmtheta(\bmx_t,t)) - \phi(\bmf_{\sg[\bmtheta]}(\bmx_{t-\Delta t},t-\Delta t))^2].
\end{align}
In other words, in contrast to standard CMs—where 
$\phi$ is fixed to $\id_{\bbR^d}$—AYT learns 
$\phi$ dynamically.
%This is in stark contrast with the standard CM, where $\phi$ is fixed rather than learned as in AYT.

% \textbf{Adversarial training for robust manifold features.} If $\bmphi$ is parametrized with a neural net, the level set relation $\phi_i(\alpha)^{-1} = \cT^i_\alpha \cM$ may not hold exactly due to brittleness of neural nets \citep{szegedy2014intriguing,goodfellow2015adversarial,nalisnick2019know}, \ie, there may be imperceptible perturbations $\bmdelta$ such that $\phi_i(\bmx+\bmdelta) = \alpha$ even though $\bmx \notin \cT^i_\alpha \cM$. To mitigate this problem, we may consider
% \begin{align}
%     \textstyle \min_{\bmphi} \max_{\|\bmdelta\|_2 \leq \tau} \bbE_{\bmx,i \in [n],\alpha \in \bbR} [ \|\phi_i(\cT^i_\alpha(\bmx+\bmdelta)) - \alpha\|_2^2] \label{eq:mf_at_loss}
% \end{align}
% instead of \eqref{eq:mf_loss} to learn manifold features. \textit{We emphasize this is distinct from adversarial fine-tuning used in, \eg, \citet{kim2024ctm}, where a discriminator is trained to distinguish data and CM samples, and the CM is trained to fool the discriminator.}

\textbf{Conceptual comparison with LPIPS.} Previous works such as \cite{song2023cm} and \cite{kim2024ctm} have used LPIPS for training CMs. Given that LPIPS also uses classifier features to define a distance between images, one may question the novelty of the manifold feature distance. Our distance distinguishes itself from LPIPS in two levels. First, our distance is tailor suited to improving CM training by aligning the tangent towards the data manifold. Second, the construction of manifold feature distance requires no human supervision and is completely self-supervised, whereas LPIPS requires ImageNet class labels and a human curated dataset of patch similarities. Furthermore, as we shall show in Section~\ref{sec:experiments}, CMs trained with manifold feature distance beats CMs trained with LPIPS, and LPIPS suffers from FID degradation possibly due to mismatch between dataset representations.

\vspace{-1mm}
\section{Experiments} \label{sec:experiments}
\vspace{-2mm}

\begin{figure}[t]
\centering
\includegraphics[width=1.0\linewidth]{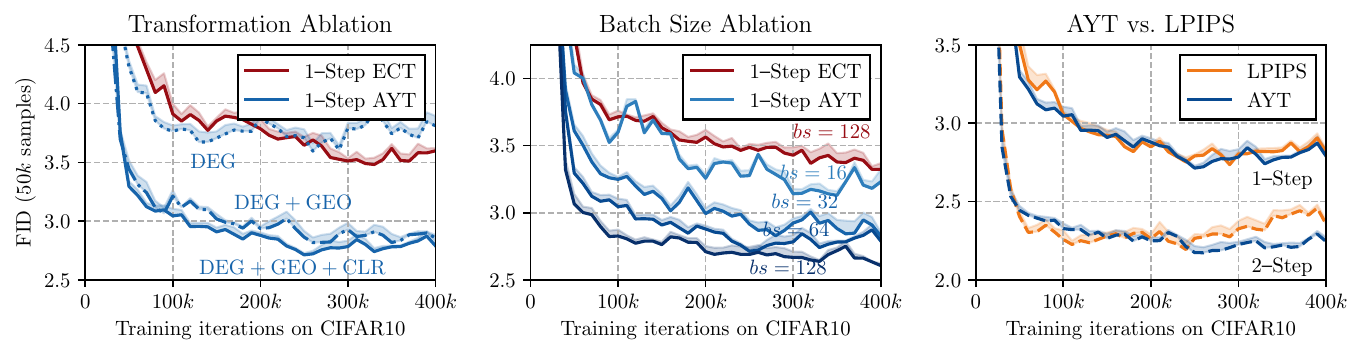}
\vspace{-6mm}
\caption{Ablation studies on CIFAR10. For transformation ablation and AYT vs. LPIPS, we use batch size $64$. Shaded regions indicate min/max FIDs over three generation trials.}
\label{fig:ablations}
\vspace{-3mm}
\end{figure}

\subsection{Ablation Studies}
\vspace{-1mm}
\textbf{Sanity check in controlled settings.} To verify whether the manifold feature distance suppresses oscillatory components in tangents, we repeated the experiments in Section~\ref{sec:analysis} with our loss in place of the mean squared error (MSE). Concretely, the tangents were now computed with \eqref{eq:feature_tangent} instead of \eqref{eq:tangent}. In the bottom row of Fig.~\ref{fig:tangent_cifar10}, we observed that the tangents were scattered and sparse, possibly implying the removal of off-manifold noise. To confirm our intuition, on the two-dimensional discs dataset, we again decomposed tangents into manifold-parallel and orthogonal components, and computed the amount of manifold-orthogonal component in the tangent. Manifold feature distance successfully removed oscillatory components from the tangent, as corroborated by the dominance of manifold-orthogonal components (right panels of Fig.~\ref{fig:discs_analysis_viz} and Fig.~\ref{fig:discs_analysis}).

\textbf{Transformation ablations.} As mentioned in Section~\ref{sec:ayt}, we considered three groups of transformations to train manifold features: three degradation-based transformations (DEG), four geometric transformations (GEO), and four color transformations (CLR). The left panel of Fig.~\ref{fig:ablations} shows changes in CM learning curves as $\bmphi$ was trained with increasing number of transformations. We observed that compounding transformations was always beneficial for CM training. Especially, the addition of geometric transformations led to the largest improvement in FID scores, implying vanilla tangents fail to provide strong training signal towards the data manifold in geometric directions.

\textbf{Robustness to batch size.} The middle panel of Fig.~\ref{fig:ablations} displays learning curves when training with batch sizes in $\{16,32,64,128\}$. Surprisingly, AYT exhibited strong FID scores even when trained with batch sizes as small as $16$, and beat ECT trained with batch size $128$. This result further affirms the oscillatory tangent hypothesis, and shows that removing oscillatory components from tangents is crucial for reducing variance during training.

\begin{wrapfigure}{r}{0.35\linewidth}
\vspace{-5mm}
\centering
\includegraphics[width=1.0\linewidth]{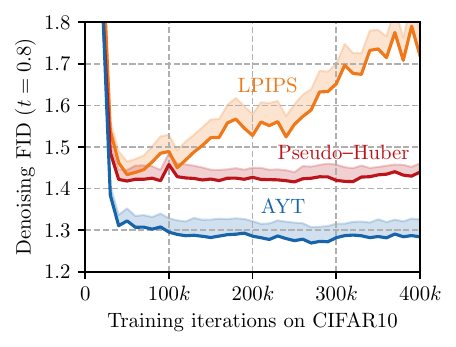}
\vspace{-7mm}
\caption{Comparison of denoising FIDs at $t = 0.8$ for CMs trained by LPIPS, pseudo-huber (PH), and manifold feature distance (AYT).}
\label{fig:dfids}
\vspace{-5mm}
\end{wrapfigure}

\textbf{AYT vs. LPIPS.} In the right panel of Fig.~\ref{fig:ablations}, we observe that AYT beats LPIPS in terms of both one and two-step generation. In particular, CM trained with LPIPS showed degradation in two-step FIDs after $200k$ iterations.
Further analysis revealed that this pathology with LPIPS was caused by inaccurate CM outputs at small $t$ corrupting outputs at larger $t$.

As shown in Fig.~\ref{fig:dfids}, denoising FIDs at $t = 0.8$ (FID between data $\bmx_0$ and denoised samples $\bmf_\bmtheta(\bmx_0+t\bmeps,t)$) for CMs trained with LPIPS diverged rapidly after $50k$ iterations.
We speculate that this behavior arises from the distributional mismatch between ImageNet and CIFAR10 (note that LPIPS is trained on ImageNet).
However, it is unclear how one can generalize LPIPS to other datasets without human supervision.
This demonstrates yet another advantage of AYT over LPIPS -- AYT presents a simple and interpretable self-supervised pipeline for constructing manifold feature distances on arbitrary datasets, enabling CM training with unbiased representations. 

\begin{table}[t]
\centering
\caption{\small{Sample quality on unconditional CIFAR10 and class-conditional ImageNet 64 $\times$ 64.}}
\label{tab:cifar_in64}
\begin{minipage}{0.45\textwidth}
\resizebox{\textwidth}{!}{%
\begin{tabular}{lcc}
\multicolumn{3}{l}{\textbf{Unconditional CIFAR10}} \\
\toprule
\textbf{Method} & \textbf{NFE} & \textbf{FID} \\
\midrule
\multicolumn{3}{l}{\textbf{Diffusion models \& Fast Samplers}} \\
DDPM & 1000 & 3.17 \\
EDM~\citep{karras2022edm} & 35 & 2.01 \\
DPM-Solver++~\citep{lu2025dpmpp} & 10 & 2.91 \\
DPM-Solver-v3~\citep{zheng2023dpm} & 10 & 2.51 \\
\midrule
\multicolumn{3}{l}{\textbf{Diffusion Distillation}} \\
DFNO (LPIPS)~\citep{zheng2023dfno} & 1 & 3.78 \\
PD~\citep{salimans2022pd} & 1 & 8.34 \\
& 2 & 5.58 \\
TRACT~\citep{berthelot2023tract} & 1 & 3.78 \\
& 2 & 3.32 \\
DMD~\citep{yin2024dmd} & 1 & 3.77 \\
SiD~\citep{zhou2024score} & 1 & 1.92 \\
CTM~\citep{kim2024ctm} & 1 & \textbf{1.87} \\
\midrule
\multicolumn{3}{l}{\textbf{Consistency Training}} \\
iCT~\citep{song2024icm} & 1 & 2.83 \\
& 2 & 2.46 \\
iCT-deep~\citep{song2024icm} & 1 & \textbf{2.51} \\
& 2 & 2.24 \\
ECT~\citep{geng2025ect} & 1 & 3.60 \\
& 2 & 2.11 \\
% ECT-fast~\citep{geng2025ect} & 1 & 3.60 \\
% & 2 & 2.11 \\
% ECT+LPIPS & 1 & \\
% & 2 & \\
ECT+AYT~(Ours) & 1 & 2.61 \\
& 2 & 2.13 \\
\bottomrule
\end{tabular} %
}
% \hspace{3mm}
\end{minipage}
\begin{minipage}{0.45\textwidth}
\resizebox{\textwidth}{!}{%
\begin{tabular}{lcc}
\multicolumn{3}{l}{\textbf{Class-Conditional ImageNet 64$\times$64}} \\
\toprule
\textbf{Method} & \textbf{NFE} & \textbf{FID} \\
\midrule
\multicolumn{3}{l}{\textbf{Diffusion models \& Fast Samplers}} \\
EDM~\citep{karras2022edm} & 79 & 2.44 \\
EDM2~\citep{karras2023edm2} & 63 & 1.33 \\
DPM-Solver~\citep{lu2022dpmsolver} & 20 & 3.42 \\
\midrule
\multicolumn{3}{l}{\textbf{Diffusion Distillation}} \\
DFNO (LPIPS)~\citep{zheng2023dfno} & 1 & 7.83 \\
PD~\citep{salimans2022pd} & 1 & 10.70 \\
 & 2 & 4.70 \\
TRACT~\citep{berthelot2023tract} & 1 & 7.43 \\
& 2 & 4.97 \\
DMD~\citep{yin2024dmd} & 1 & 2.62 \\
DMD2~\citep{yin2024dmd2} & 1 & \textbf{1.28} \\
SiD~\citep{zhou2024score} & 1 & 1.52 \\
CTM~\citep{kim2024ctm} & 1 & 1.92 \\
& 2 & 1.73 \\
% Moment Matching~\citep{salimans2024multistepmoment} & 1 & 3.00 \\
% & 2 & 3.86 \\
\midrule
\multicolumn{3}{l}{\textbf{Consistency Training}} \\
iCT~\citep{song2024icm} & 1 & 4.02 \\
& 2 & 3.20 \\
iCT-deep~\citep{song2024icm} & 1 & 3.25 \\
& 2 & 2.77 \\
ECT-S~\citep{geng2025ect} & 1 & 5.51 \\
 & 2 & 3.18 \\
ECT-S+AYT (Ours) & 1 & 4.42 \\
 & 2 & 3.27 \\
\bottomrule
\end{tabular} %
}
\end{minipage}
\vspace{-.2in}
\end{table}

\vspace{-2mm}
\subsection{Comparison with other methods.} 
\vspace{-2mm}
To evaluate the effectiveness of our method, we compare against three major families of few-step generative approaches: 
advanced diffusion samplers,
consistency models, and 
diffusion distillation methods.
We report FID scores across methods and numbers of function evaluations (NFE) in Tab.~\ref{tab:cifar_in64}.

\textbf{Comparison within consistency models.}
%\vspace{-2mm}
On CIFAR10, AYT improves the 1-step FID from $3.60$ to $2.61$ over Easy Consistency Training (ECT), while maintaining comparable 2-step performance ($2.11$ vs. $2.13$). Notably, AYT also outperforms Improved Consistency Training (iCT) ($2.83$ FID), despite the latter relying on multi-stage training schedules that progressively reduce timestep gaps. On ImageNet $64 \times 64$, AYT outperforms ECT by a non-trivial margin in both 1- and 2-step settings, reducing 1-step FID from $5.51$ to $4.41$, and maintains competitive 2-step performance ($3.27$ vs. $3.18$). It also achieves competitive performance relative to iCT, while using significantly fewer resources—most notably, a batch size of 128, which is $8\times$ smaller than the 1024 used by iCT. These results highlight the effectiveness of our tangent alignment strategy in stabilizing consistency model training, without the need for schedule tuning, multi-stage optimization, or large-scale training infrastructure.

\textbf{Comparison with distillation.}
On CIFAR10, our method achieves competitive performance compared to state-of-the art diffusion distillation models such as Consistency Trajectory Model (CTM, FID $1.87$) and Score Identity Distillation (SiD, FID $1.92$), despite not relying on any pretrained teacher model or adversarial training. On Imagenet $64 \times 64$, AYT outperforms several distillation-based approaches while reducing the gap between state-of-the-art distillation approaches.
This result is particularly notable given that these baselines often inherit strong priors and score functions from large pretrained diffusion models.
In contrast, we train our model from scratch, yet reach comparable or superior sample quality.

\textbf{Comparison with fast samplers.}
Finally, we compare with high-order diffusion ODE solvers.
Our 2-step performance surpasses that of methods such as DPM-Solver++ and DPM-Solver-v3 that operate with NFE $\geq 10$, despite our significantly smaller sampling cost.
These results affirm the practicality of our method for high-quality generation under extreme computational budgets.

\section{Discussion}
\vspace{-2mm}
\textbf{Further implications.} 
Beyond images, it will be interesting to explore other domains such as audio, text, or multimodal data with diverse augmentation strategies. For example, in audio data, common augmentations include time-stretching, pitch-shifting, masking, or adding background noise, all of which can be utilized for learning data manifold features. Applying our approach in such settings could provide valuable insights into how well the proposed distance metric generalizes across modalities. Such extensions could further demonstrate the generality of the framework.

\textbf{Limitations.} Our study has so far focused on relatively small-scale settings. While the method requires additional training and increases memory usage, the auxiliary classifier is lightweight: it trains much faster than the main model and adds little memory overhead. As a result, we expect these constraints to be less critical in practice, even when scaling to larger datasets.
We have also restricted our evaluation up to resolution $64 \times 64$. While higher-resolution experiments remain open, the consistent improvements on CIFAR10 and ImageNet suggest that the approach may transfer well to more demanding settings. Moreover, recent high-resolution training often relies on latent diffusion models (LDMs)~\citep{rombach2022ldm}, which downsample images by a factor of 8, making our method potentially well-suited for such pipelines.
In this sense, the present work should be viewed as a first step: AYT establishes strong evidence on standard benchmarks while opening several promising directions for scaling and broader applications in self-supervised and generative learning.

\section{Conclusion}
\vspace{-2mm}
In this paper, we analyzed the training dynamics of consistency models (CMs) and showed that their update directions (tangents) often contain manifold-parallel oscillatory components, which slow convergence. Motivated by this, we introduced the MFD -- a simple, self-supervised objective computed in the feature space of an auxiliary network trained to be sensitive to off-manifold perturbations. By aligning tangents toward the data manifold (i.e., amplifying manifold-orthogonal components), MFD contracts CM trajectories more efficiently without relying on human supervision or curated perceptual datasets. Empirically, MFD stabilizes training by orders of magnitude over the pseudo-Huber loss while improving sample quality. On CIFAR10 and class-conditional ImageNet $64 \times 64$, our method outperforms consistency-training baselines, attains FIDs competitive with distillation approaches despite training from scratch, and remains robust even with very small batch sizes (e.g., 16). These results indicate that matching the optimization geometry of CMs to the structure of the data manifold is a practical and powerful route to faster, more reliable few-step generation.

\newpage

% \section*{Ethics Statement}

% \section*{Reproducibility Statement}

% We describe all experimental procedures in Appendix~\ref{append:settings}. Code will be published via GitHub if the paper is accepted, and a link to the GitHub project will be included in the camera ready version.



{\small
\bibliography{iclr2025_conference}
\bibliographystyle{iclr2025_conference}
}

\newpage

\appendix

\section{Experiment Settings} \label{append:settings}

We build our method on top of the Easy Consistency Training (ECT)~\citep{geng2025ect} framework with minor modifications. 
Unless otherwise noted, we follow ECT defaults for data preprocessing, forward process $\bmx_t = \bmx + t\bmeps$, timestep sampling, and evaluation protocol.

\subsection{Data Preprocessing}

CIFAR10 and ImageNet $64 \times 64$ datasets are preprocessed with code provided by \cite{karras2022edm} in \url{https://github.com/NVlabs/edm}.

\subsection{Model Architectures and Initialization}
\label{append:model}

\textbf{Classifier architectures.} We use VGG16 classification networks~\citep{simonyan2015vgg} to parametrize manifold features. All VGG16 networks are trained from scratch without any special initialization schemes.

\textbf{Consistency model architectures.} We adopt the same backbone choices as ECT. Specifically, we use DDPM++~\citep{song2021sde} for CIFAR10 and EDM2-S~\citep{karras2023edm2} for ImageNet $64 \times 64$. On both datasets, we initialize the consistency model with a pretrained diffusion model of the corresponding architecture.

\subsection{Classifier Training Configurations}

We use identical training configurations on CIFAR10 and ImageNet $64 \times 64$. Specifically, we use the Adam optimizer \citep{kingma2015adam} with learning rate $0.0001$ and batch size $512$. Each manifold feature is trained for $400k$ iterations to minimize \eqref{eq:mf_loss}. Our color and geometric transformation pipeline largely follows that described in Appendix B of \citet{karras2020ada}. We describe the augmentation pipeline for degradations below. Specifically, given $\bmx \sim p(\bmx)$,
\begin{itemize}
    \item \textbf{Gaussian noise.} Sample $\alpha \sim \unif(0,\alpha_{\max})$, $\bmeps \sim \cN(\bm{0},\bm{I})$, return $\bmx + \alpha \bmeps$.
    \item \textbf{Gaussian blur.} Sample $\alpha \sim \unif(0,\alpha_{\max})$ , return $\bmkappa_\alpha \circledast \bmx$.
    \item \textbf{Mixup.} Sample $\alpha \sim \unif(0,0.5)$ and another datapoint $\bmy$, return $(1-\alpha) \bmx + \alpha \bmy$.
\end{itemize}
Here, $\bmkappa_\alpha$ is a Gaussian blur kernel with sigma $\alpha$.

\subsection{Consistency Model Training Configurations}
\label{append:train}

We use a global batch size of 128 for all runs, except in the batch-size ablation. Exponential moving average (EMA) is enabled throughout training, with dataset-specific settings detailed below.

\textbf{CIFAR10.}
We train consistency models for 400K iterations without any multi-stage schedule unlike iCT/ECT. We use the RAdam optimizer \citep{liu2020radam} with learning rate $0.0001$ and exponential moving average (EMA) decay rate of $0.9999$.

\textbf{ImageNet $64 \times 64$.}
We train for $200k$ iterations with the same multi-stage schedule as ECT. To mitigate early-stage overfitting, we enable our loss (AYT) after 75K iterations.
We use the Adam optimizer \citep{kingma2015adam} with learning rate $0.001$ and an inverse-square-root decay schedule of decay parameter $2000$. EMA follows the Power-EMA formulation introduced in EDM2, but we do not apply post-hoc EMA after training.

\subsection{Sampling and Evaluation}
\label{append:eval}

We evaluate 1-step and 2-step sampling with Fréchet Inception Distance (FID), computed between the training set and 50K generated samples. For 2-step sampling, the intermediate timestep is fixed to $t=0.821$ on CIFAR10 and $t=1.526$ on ImageNet $64 \times 64$, following ECT. Unless otherwise stated, we follow the ECT evaluation setup and report FID computed with $50k$ samples.

% \begin{equation}
%     \frac{d\bmphi}{dt}^\top \bm{J}_{\bmphi}(\bmf_{\bmtheta}(\bmx_t,t))
% \end{equation}

\newpage

\section{Consistency Model Samples}

\begin{figure}[H]
\centering
\begin{subfigure}{1.0\linewidth}
\includegraphics[width=1.0\linewidth]{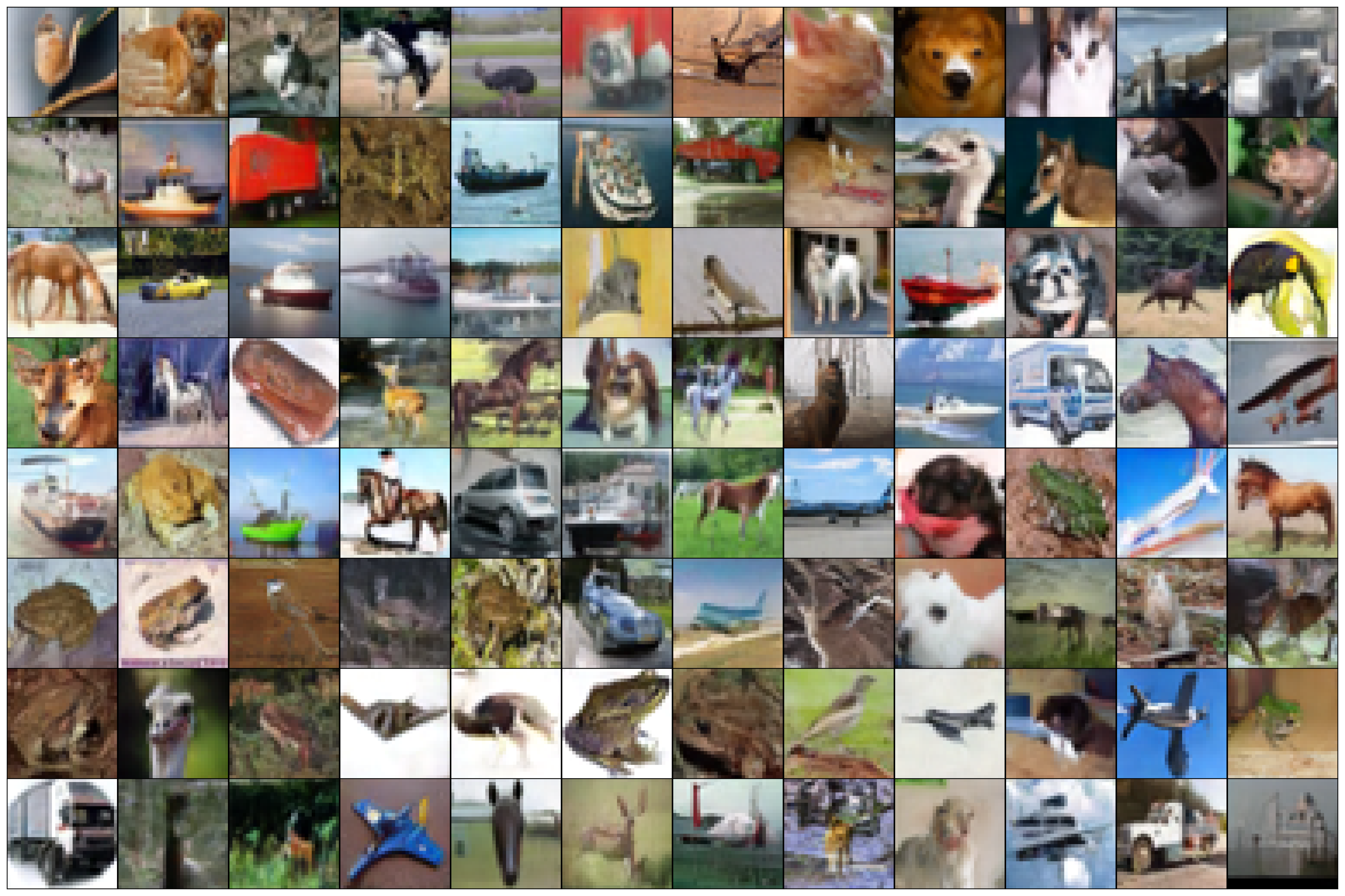}
\caption{Samples from CM trained via ECT.}
\vspace{4mm}
\end{subfigure}
\begin{subfigure}{1.0\linewidth}
\includegraphics[width=1.0\linewidth]{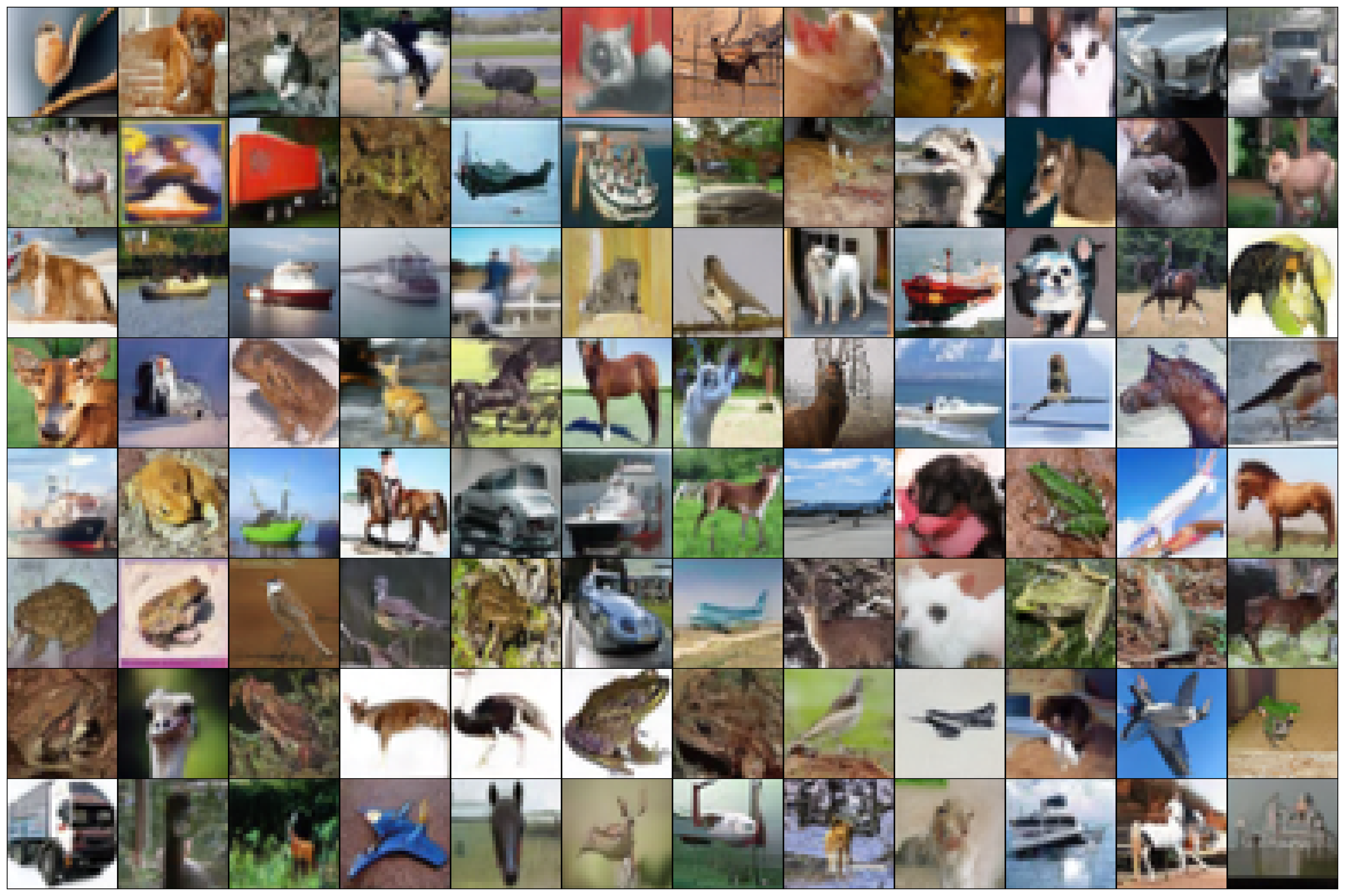}
\caption{Samples from CM trained via AYT (Ours).}
\end{subfigure}
\caption{Uncurated one-step CM samples on CIFAR10.}
\label{fig:cifar10_samples}
\end{figure}

\newpage

\begin{figure}[H]
\vspace{10mm}
\centering
\begin{subfigure}{1.0\linewidth}
\includegraphics[width=1.0\linewidth]{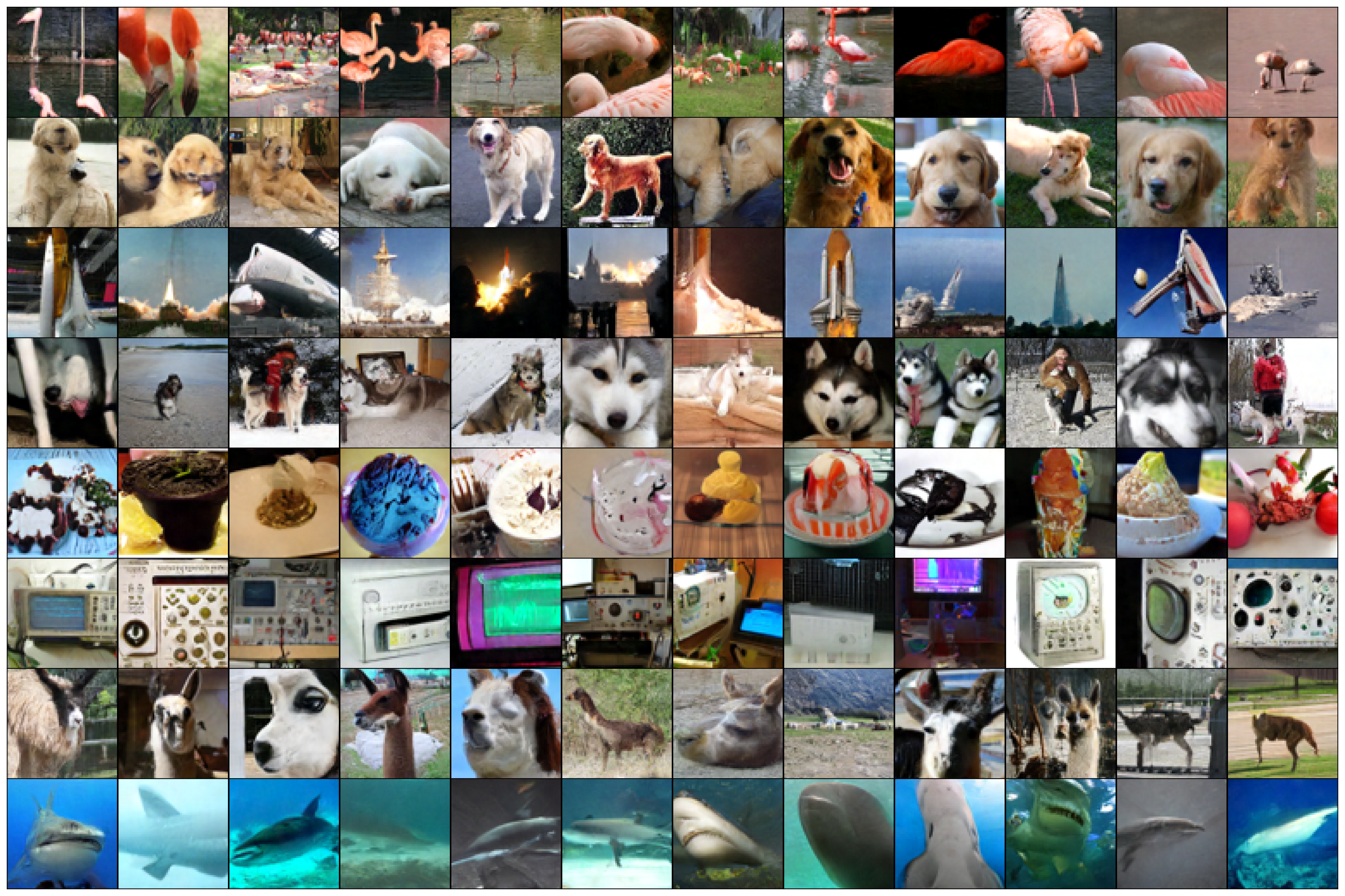}
\caption{Samples from CM trained via ECT.}
\vspace{4mm}
\end{subfigure}
\begin{subfigure}{1.0\linewidth}
\includegraphics[width=1.0\linewidth]{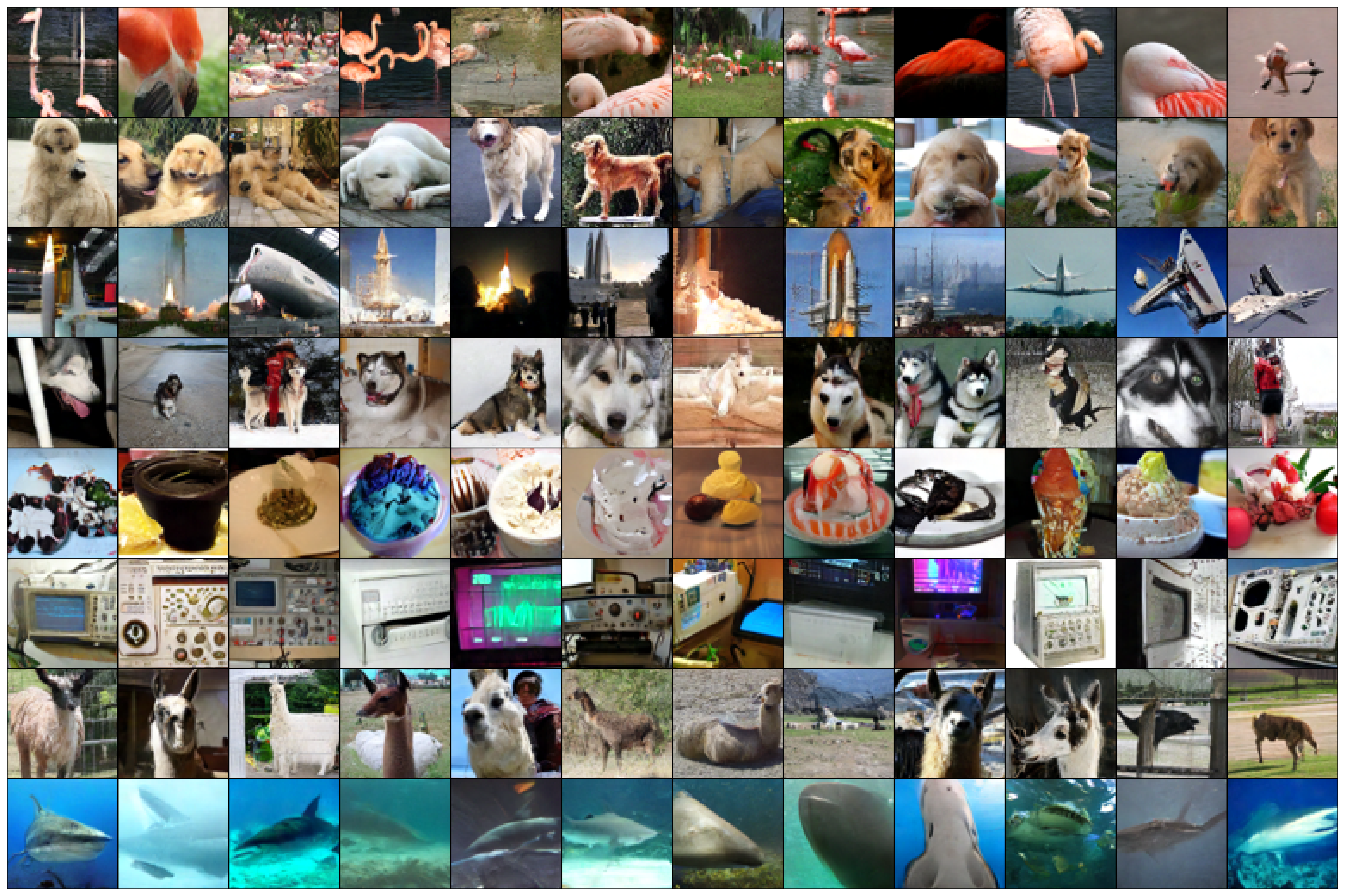}
\caption{Samples from CM trained via AYT (Ours).}
\end{subfigure}
\caption{Uncurated one-step CM samples on ImageNet $64 \times 64$.}
\label{fig:imagenet_samples}
\end{figure}

\end{document}